\documentclass[10pt,twocolumn,letterpaper]{article}

\usepackage[pagenumbers]{iccv} 

\definecolor{iccvblue}{rgb}{0.21,0.49,0.74}
\usepackage[pagebackref,breaklinks,colorlinks,allcolors=iccvblue]{hyperref}
\usepackage{multirow}
\usepackage{colortbl}

\title{SignRep: Enhancing Self-Supervised Sign Representations}

\author{Ryan Wong$^1$, Necati Cihan Camgoz$^2$, Richard Bowden$^1$ \\
$^1$University of Surrey, 
$^2$Meta Reality Labs\\
\texttt {\{r.wong, r.bowden\}@surrey.ac.uk, neccam@meta.com}
}

\begin{document}
\maketitle
\begin{abstract}

Sign language representation learning presents unique challenges due to the complex spatio-temporal nature of signs and the scarcity of labeled datasets. Existing methods often rely either on models pre-trained on general visual tasks, that lack sign-specific features, or use complex multimodal and multi-branch architectures. To bridge this gap, we introduce a scalable, self-supervised framework for sign representation learning. We leverage important inductive (sign) priors during the training of our RGB model. To do this, we leverage simple but important cues based on skeletons while pretraining a masked autoencoder. These sign specific priors alongside feature regularization and an adversarial style agnostic loss provide a powerful backbone. Notably, our model does not require skeletal keypoints during inference, avoiding the limitations of keypoint-based models during downstream tasks. When finetuned, we achieve state-of-the-art performance for sign recognition on the WLASL, ASL-Citizen and NMFs-CSL datasets, using a simpler architecture and with only a single-modality. Beyond recognition, our frozen model excels in sign dictionary retrieval and sign translation, surpassing standard MAE pretraining and skeletal-based representations in retrieval. It also reduces computational costs for training existing sign translation models while maintaining strong performance on Phoenix2014T, CSL-Daily and How2Sign.

\end{abstract}    
\section{Introduction}
\label{sec:intro}

Sign language is an important means of communication for millions of people worldwide.
Sign languages have complex visual characteristics, which include intricate hand shapes, motions, body poses and facial expressions that models need to accurately interpret and process \cite{braem2001hands}. Furthermore, the computational demands of processing long video sequences add considerable challenges, making it difficult to scale these systems effectively. As a result, current approaches to sign recognition and translation often rely on general pretrained vision models \cite{albanie2020bsl,li2020word,camgoz2020sign,hu2021global,jiang2021sign,zuo2023natural,wongsign2gpt}. 
Creating robust, label-free sign language representations that generalize across diverse datasets is challenging, yet essential for scalable sign language modeling.

Most existing methods for sign recognition rely on multi-modal or an ensemble of specialized models to achieve state-of-the-art recognition results. These models often require multi-channel inputs (e.g. RGB, depth and skeleton data) or specialized architectures to capture the complex interactions of hand, body or facial expressions \cite{jiang2021skeleton,jiang2021sign,zuo2023natural,hu2021signbert,hu2023signbert+,zhao2023best}. With each country having its own sign language and linguistic study expensive, currently available labeled datasets are sparse with typically under 2000 unique signs \cite{li2020word,sincan2020autsl,albanie2020bsl,hu2021global,joze2018ms}. 
Collecting annotated sign language data is costly and time-consuming, making it infeasible to rely solely on labeled datasets for sign representation learning. This highlights the need for methods that can learn sign language representations from large-scale, unlabeled data.

Newer methods for pretraining sign language translation models often require sentence-aligned annotations \cite{wongsign2gpt,zhou2023gloss,zhou2024scaling,rust2024towards,jiao2025visual}. While these prior approaches jointly learn individual sign representations and inter-sign relationships, they are not scalable to unlabeled datasets as they require sentence-aligned annotations. We therefore focus on building a scalable and generalizable individual sign representation framework using self-supervised learning. A strong foundation in individual sign embeddings is an important step before building effective inter-sign models, as it ensures robust feature representations before incorporating complex long range temporal information.

Our main contributions are as follows:
(1) We propose a scalable self-supervised Masked Autoencoding (MAE) framework which leverages sign priors, adversarial loss and feature regularizations for sign representation learning.
(2) We introduce a method for analyzing sign class similarities in unseen datasets, introducing an auxiliary class probability distribution loss, which enhances recognition performance.
(3) Our single pretrained model achieves state-of-the-art sign recognition, surpassing complex architecture and multimodal models.
(4) We demonstrate the effectiveness of our sign representations for sign dictionary retrieval, achieving strong performance without any downstream training.
(5) We demonstrate that using our pretrained model as a frozen feature extractor, training is more tractable by reducing memory requirements and improving the performance of existing sign translation models.

\section{Related Work}
\label{sec:rw}

\noindent\textbf{Supervised Sign Recognition.}
Isolated Sign Recognition involves identifying a single sign within a given video.
Sign recognition models often rely on pretrained spatio-temporal models from action recognition datasets such as Kinetics \cite{carreira2017quo}, fine-tuned for specific sign recognition tasks \cite{albanie2020bsl,li2020word,hu2021global,jiang2021sign,zuo2023natural}. This transfer learning approach faces a domain shift problem, as sign videos have unique temporal dynamics and subtle gestures not well-represented in general action datasets, leading to performance degradation.
To leverage invariance, some methods use skeletal keypoints instead of full-frame RGB models \cite{dafnis2022bidirectional,wong2023learnt,hu2021signbert,hu2023signbert+,zhao2023best}. While keypoint-based models are more memory-efficient, they typically underperform relative to RGB models \cite{albanie2021bbc,desai2024asl,raude2024tale} and require complex architectural modifications to Graph Convolutional Networks (GCNs) \cite{liu2020disentangling} or transformers \cite{vaswani2017attention}, to model the spatial-temporal relationships. Moreover, keypoints are prone to errors, such as missing or misdetected points \cite{moryossef2021evaluating,moryossef2024optimizing}.
To solve these issues, recent state-of-the-art methods have combined keypoint and RGB modalities through branching or ensembling techniques \cite{jiang2021sign,jiang2021skeleton,zuo2023natural}. While they achieve higher accuracy, they increase computational complexity due to the need for multiple models or support for multi-modal inputs.\\

\noindent\textbf{Self-Supervised Sign Representation Learning.} Self-supervised learning for vision \cite{oquab2023dinov2,tong2022videomae,ryali2023hiera,assran2023self} and language \cite{devlin2018bert,radford2019language} has demonstrated substantial benefits by leveraging unlabeled datasets for various tasks. In sign language, approaches such as Skeletor \cite{jiang2021skeletor}, SignBERT \cite{hu2021signbert}, SignBERT+ \cite{hu2023signbert+} and BEST \cite{zhao2023best} focus on keypoint-based self-supervised learning, employing masked learning similar to BERT \cite{devlin2018bert}. However, these methods depend on keypoints as inputs and therefore requiring ensembling with RGB models to achieve competitive performance, resulting in increasing complexity and limited pretraining benefits. This highlights the need for a simple and efficient model which capturing sign-specific information and retains the pretraining advantages.\\

\noindent\textbf{Paired Sign-Text Pretraining for Sign Translation.}
Supervised pretraining approaches for sign translation, such as GFSLT-VLP \cite{zhou2023gloss}, Sign2GPT \cite{wongsign2gpt}, SignHiera \cite{rust2024towards}, MSLU \cite{zhou2024scaling} and VAP \cite{jiao2025visual}, use sign video–spoken language pairs to enhance sign translation. Sign2GPT employs a pretrained DinoV2 \cite{oquab2023dinov2}, which is effective for general vision tasks but requires LoRA \cite{hu2022lora} fine-tuning for sign translation to achieve strong performance, significantly increasing computational costs. This highlights the domain gap between pretrained foundation vision models and sign language.
SignHiera \cite{rust2024towards} demonstrates success with large-scale training on YT-ASL \cite{uthus2024youtube} using Hiera MAE \cite{ryali2023hiera} with language-supervised pretraining. However, they acknowledge that SignHiera requires significant computational resources (64 A100 GPUs for two weeks), emphasizing the need for more accessible and cost-effective video pretraining on large-scale datasets. 
These methods also face scalability challenges due to their reliance on sentence-aligned annotations. In this paper, we introduce a scalable self-supervised pretraining framework that learns sign representations without paired sign-text annotations, aiming to significantly reduce computational costs for pretraining and improve efficiency in downstream sign tasks.

\section{Sign Priors}

\begin{figure*}[t]
    \centering
    \includegraphics[width=1.0\linewidth]{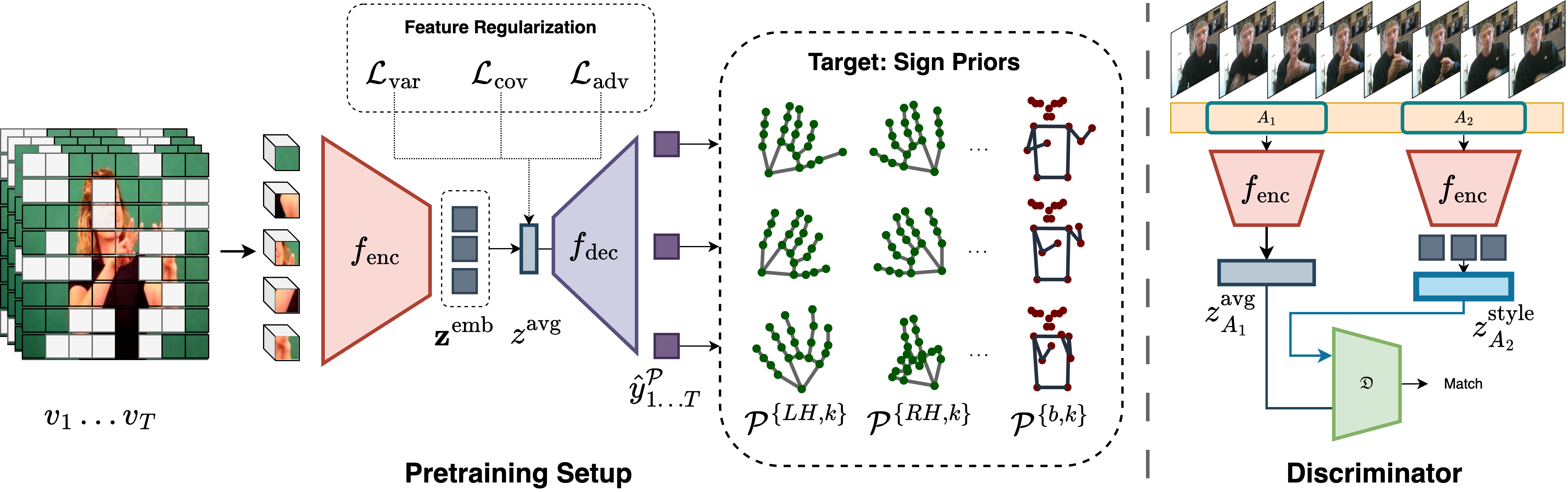}
    \caption{(Left): The pretraining process for SignRep, which leverages masked representation learning to predict sign priors such as hand keypoints and joint angles. This is achieved through a Hiera encoder and a lightweight sign decoder. The representation is further refined with regularization losses, including variance, covariance and adversarial style loss. (Right): An example setup for the discriminator to obtain a representation pair to predict a style-representation match.}
    \label{fig:main}
\vspace{-1.0em}
\end{figure*}

Sign language communication relies on hand shapes, body posture and interactions. To guide the model toward meaningful sign representations, we introduce ``sign priors'', a set of cues that we know capture essential sign features. These priors ($\mathcal{P}$) serve as the primary targets for our model, as detailed in \cref{ss:sign_prior_recon}. Using a human pose estimation model specialized for sign language videos \cite{ivashechkin2023improving}, we extract joint angles and 3D keypoints. We categorize our priors into \textbf{keypoint}, \textbf{angle}, \textbf{distance} and \textbf{signer activity}.\\

\noindent\textbf{Keypoint Priors} define the spatial structure of signing, we divide this into two priors:

\textit{Hand Keypoint Prior ($\mathcal{P}^{\{h, \text{k}\}}$):} This prior is designed to capture hand shapes and orientations. We normalize the 3D coordinates of the 21 hand keypoints by setting the wrist as the origin, thereby eliminating positional variations in the body space. The resulting vector, $\mathcal{P}^{\{h, \text{k}\}} \in \mathbb{R}^{21 \times 3}$, represents the configuration of the left or right hand, where  $h \in \{ \text{LH}, \text{RH} \}$ denotes the left or right hands.

\textit{Full Body Keypoint Prior (\( \mathcal{P}^{\{b, \text{k}\}} \)).} This prior captures the positioning of the body and hands within the body space. We use all 61 3D keypoints (21 for each hand and 19 for the body), resulting in \( \mathcal{P}^{\{b, \text{k}\}} \in \mathbb{R}^{61 \times 3} \).\\

\noindent\textbf{Joint Angle Priors} capture finger flexion, extension and encode upper-body posture variations:

\textit{Hand Joint Angle Prior (\( \mathcal{P}^{\{h, \text{a}\}} \)).} This prior captures hand joint orientations. We extract 41 hand joint angles, yielding a vector \( \mathbb{R}^{41 \times 1} \). To handle the continuous nature of angles, we apply sine and cosine transformations, resulting in the final angle prior \( \mathcal{P}^{\{h, \text{a}\}} \in \mathbb{R}^{41 \times 2} \).

\textit{Body Angle Pior (\( \mathcal{P}^{\{b, \text{a}\}} \)).} To capture the body's joint orientations, we extract the 22 body joint angles. These angles are transformed using sine and cosine functions to handle their continuous nature, resulting in the body angle prior \( \mathcal{P}^{\{b, \text{a}\}} \in \mathbb{R}^{22 \times 2} \).\\

\noindent\textbf{Distance Priors} capture fine-grained differences between signs and the interactions of hands in the signing space:

\textit{Fingertip Distance Prior ($\mathcal{P}^{\{h, \text{d}\}}$).} Since fingertip interactions are a key component in sign language, we compute a distance matrix by measuring the distances from the fingertip keypoints to each of the knuckle, wrist and other fingertip keypoints. This produces the prior \( \mathcal{P}^{\{h, \text{d}\}} \in \mathbb{R}^{5 \times 11 \times 3} \).

\textit{Hand-Interaction Distance Prior (\( \mathcal{P}^{\{b, \text{d}\}} \)).} Similar to the fingertip distance prior, we calculate a distance matrix using the wrist and five fingertip keypoints from each hand, resulting in a \( 6 \times 2 \) set of keypoints. The distance between the hands, body and facial keypoints is computed to capture the interactions between the hands and other body parts, resulting in the final hand-interaction prior \( \mathcal{P}^{\{b, \text{d}\}} \in \mathbb{R}^{12 \times 22 \times 3} \).\\

\noindent\textbf{Signer Activity.} Some signs require only one hand or the signer may be in a resting pose. To enable the model to capture this information, we develop a simple heuristic. We define the active prior \( \mathcal{P}^{\{h,\texttt{act}\}}\), where a hand is considered inactive if it remains below the middle of the stomach and has not moved over the course of a video clip. This results in a prior \( \mathcal{P}^{\{h,\texttt{act}\}} \in [0,1] \), indicating whether each hand is active (1) or inactive (0).

\section{SignRep Architecture}

To learn the sign priors, we introduce a novel adaptation of the Hierarchical Vision Transformer (Hiera) \cite{ryali2023hiera}, designed with a pretraining task specifically tailored for sign representation learning without requiring labeled sign data. We choose the Hiera model as it aligns with our objective of being efficient, simple and effective. It has demonstrated strong performance in masked representation learning while eliminating the need for complex architectures and specialized modules. We aim to develop a single spatio-temporal sign model that enhances performance without relying on multi-branch or ensemble methods, instead improving the representation learning process.

The standard Hiera MAE framework processes a video clip represented as $V = \{v_1, v_2, \dots, v_t, \dots, v_T\}$, where  $v_t$  denotes the frame at time step $t$, and $T$ is the total number of frames. The video is first divided into spatiotemporal patches. A masking strategy is then applied, with a subset of patches randomly masked according to a masking ratio $M$. The unmasked patches are fed into the encoder $f_{\text{enc}}$, which processes them to generate an output representation $\mathbf{z}^{\text{emb}} = \{z^{\text{emb}}_0, z^{\text{emb}}_1, \dots, z^{\text{emb}}_K\}$, where  $K$  is the number of output tokens from the encoder. 
The representation $\mathbf{z}^{\text{emb}}$ is then passed to a decoder, along with learnable masked tokens as input. The pretraining objective is to reconstruct the masked patches through pixel-level prediction, allowing the model to learn contextual relationships and enhance the visual representation. 

Rather than pixel reconstruction, we use sign priors to learn meaningful sign representations, replacing the pixel reconstruction decoder with a lightweight sign decoder.

\paragraph{Sign Decoder ($f_{dec}$).}
As shown in \cref{fig:main}, instead of passing $\mathbf{z}^{\text{emb}}$ to the decoder, we take the average across $\mathbf{z}^{\text{emb}}$ tokens, which is then processed through layer normalization followed by a fully connected layer. This produces an output representation \( z^{\text{avg}} \in \mathbb{R}^{1\times D} \), where \( D \) is the dimensionality of the representation.
The representation \( z^{\text{avg}} \) is then upsampled temporally to match the sequence length \( T \) of the input video.
The upsampling module is implemented using a lightweight network consisting of a 1D convolution with a kernel size of 1, followed by a GELU activation function, and then a transpose convolution with a kernel size of \( T \) to match the number of input frames and an output dimension of \( D' \). This results in an upsampled vector \( \mathbf{z}^{\text{up}} \in \mathbb{R}^{T \times D'} \), where $\mathbf{z}^{up}=[{z^{up}_1, z^{up}_2, \dots, z^{up}_T}]$.

We can then incorporate prediction heads for each sign prior to the network as follows:
\begin{equation}
\hat{y}^{\mathcal{P}}_{t} = {g}_{\mathcal{P}}({z}^{\text{up}}_t)
\end{equation}
where each sign prior,
    $\mathcal{P}\in\{\mathcal{P}^{\{h,k\}}, \allowbreak \mathcal{P}^{\{h,a\}}, \allowbreak \mathcal{P}^{\{h,d\}}, \allowbreak \mathcal{P}^{\{b,k\}}, \allowbreak \mathcal{P}^{\{b,a\}},  \allowbreak \mathcal{P}^{\{b,d\}}\}$
has a corresponding fully connected layer ($g_{\mathcal{P}}$) with an associated output \( \hat{y}^{\mathcal{P}}_{t} \) which matches the flattened dimension of the corresponding sign prior.
For the activity prior we simply use an MLP layer using $z^{\text{avg}}$ as input which produces an output of 2 to match the dimension of $[\mathcal{P}^{\{\text{LH},\text{act}\}},\mathcal{P}^{\{\text{RH},\text{act}\}}]$.

The decoder is completely removed during downstream tasks and only the encoder is used as the sign representation model. As a result, the target sign priors are not required during downstream tasks, allowing the use of a single model that inherently captures sign knowledge. This approach ensures computational efficiency by leveraging masked representation learning while directly learning the essential sign priors.

\section{Sign Representation Objectives}

\subsection{Sign Priors Reconstruction}
\label{ss:sign_prior_recon}

In the previous section, we identified the sign priors that our model needs to learn. 
Since the sign decoder applies upsampling to match the number of input frames, each frame's prior has an associated output prediction. For each predicted sign prior output, $\hat{y}^{\mathcal{P}}_t$, we train the model to regress to the corresponding target value $y^{\mathcal{P}}_t$ from prior $\mathcal{P}$ at frame $t$, using smooth L1 loss ($L1$) across all sign priors for each frame in the input video sequence. The same loss is applied to the sign activity prior for consistency. 
We also mitigate the impact of low-quality keypoints in our objective function by only using keypoints with over 50\% confidence from the pose estimation model and masking missing keypoints from the loss. 
The final reconstruction loss is as follows:
\begin{equation}
    \mathcal{L}_{\text{recon}} = \sum_{\mathcal{P} \in \mathbf{P}} w_{\mathcal{P}} L1(\hat{y}^{\mathcal{P}}, y^{\mathcal{P}})
\end{equation}
where $w_{\mathcal{P}}$ is the loss weighting for the prior $\mathcal{P}$.

\subsection{Representation Regularization}

\paragraph{Feature Regularization.}

Self-supervised methods have shown that regularizing features improves representation quality \cite{bardes2022vicreg,zbontar2021barlow}. To enhance our representations, we incorporate a variance and covariance loss into $z^{\text{avg}}$.

The \textbf{variance loss}, $\mathcal{L}_{\text{var}}$, encourages diversity in the learned representations by spreading them across the representation space. To compute the variance loss, we first calculate the standard deviation for each feature dimension ($j$) across a batch ($N$).
\begin{equation}
    \sigma_j = \sqrt{\frac{1}{N - 1} \sum_{i=1}^{N} (z^{\text{avg}}_{i,j} - \bar{z}^{\text{avg}}_j)^2}    
\end{equation}
The diversity of the features are then encouraged through a hinge function to ensure that the variance does not fall below a threshold of one.
\begin{equation}
    \mathcal{L}_{\text{var}} = \sum_{j=1}^{D} \max(0, 1 - \sigma_j)    
\end{equation}

The \textbf{covariance loss}, $\mathcal{L}_{\text{cov}}$, reduces correlations among features, helping to avoid redundancy. To compute the covariance loss, we first need to calculate the covariance matrix such that:
\begin{equation}
    \mathcal{C}_{j,k} = \frac{1}{N} \sum_{i=1}^{N} (z^{\text{avg}}_{i,j} - \bar{z}^{\text{avg}}_j)(z^{\text{avg}}_{i,k} - \bar{z}^{\text{avg}}_k)
\end{equation}
where $\mathcal{C}_{j,k}$ represents the covariance between feature dimensions $j$ and $k$ and $\bar{z}^{\text{avg}}_j$ is the mean of the $j$-th feature across the batch.
The loss then penalises off-diagonal values in the covariance matrix, encouraging different feature dimensions to be uncorrelated:
\begin{equation}
    \mathcal{L}_{\text{cov}} = \sum_{j \neq k} \mathcal{C}_{j,k}^2    
\end{equation}

\paragraph{Style Agnostic Representations.}
\label{para:style}

We also aim to encourage the encoder to capture robust and generalizable sign features, while filtering out irrelevant details such as background and person-specific appearance. We can explicitly learn this by introducing a discriminator that evaluates whether two representations share the same ``style'', defined here by background and appearance features.

To extract style information, we calculate the gram matrix, commonly used in image style transfer \cite{gatys2016image}, from the $\mathbf{z}^{\text{emb}}$ tokens, averaging across the column dimension to produce the style representation  $z^{\text{style}}\in \mathbb{R}^D$. We then pair $z^{\text{avg}}$ with $z^{\text{style}}$ and pass them to the discriminator, which learns to output 1 for matching pairs and 0 for non-matching pairs.

To generate these representations pairs, as shown in \cref{fig:main}(right), we randomly crop a video sequence of length $L$ to create two segments, $A_1$ and $A_2$, each of length $T$. For each, we extract $z^{\text{avg}}_{A_1}$ and  $z^{\text{style}}_{A_1}$, as well as $z^{\text{avg}}_{A_2}$ and  $z^{\text{style}}_{A_2}$, assuming that they share background and appearance features due to originating from the same video. From a different video, we obtain segment $B$ with its style $z^{\text{style}}_{B}$, which provides a contrasting background and appearance.

The discriminator, $\mathfrak{D}$, is then trained to produce 0 output for mismatched styles and 1 for matched styles, such that $\mathfrak{D}(z^{\text{avg}}_{A_1}, z^{\text{style}}_{B}) = 0$ and  $\mathfrak{D}(z^{\text{avg}}_{A_1}, z^{\text{style}}_{A_2}) = 1$. This can be used to further enhance the generalization of the encoder, we add an adversarial loss during pretraining, designed to fool the discriminator which encourages the model to focus on sign-specific content over style-related features. The adversarial loss is formalized as follows:
\begin{align}
    \mathcal{L}^{(A_1,A_2)}_{pos} &= \max(0,\mathfrak{D}(z^{\text{avg}}_{A_1}, z^{\text{style}}_{A_2})-\mathbb{E}_{q\sim \text{U}}\mathfrak{D}(q)) \\
    \mathcal{L}^{(A_1,B)}_{neg} &= \max(0,\mathbb{E}_{q\sim \text{M}}\mathfrak{D}(q)-\mathfrak{D}(z^{\text{avg}}_{A_1}, z^{\text{style}}_{B}))  \\
    \mathcal{L}^{A_1}_{\text{adv}} &= (\mathcal{L}^{(A_1,A_2)}_{pos})^2 + (\mathcal{L}^{(A_1,B)}_{neg})^2
    \label{eq:disc}
\end{align}
where $\mathbb{E}_{q\sim \text{M}}\mathfrak{D}(q)$ is the expected discriminator output of style for representation pairs that come from the same video sequence and $\mathbb{E}_{q\sim \text{U}}\mathfrak{D}(q)$ for when they come from different video sequence. We apply this loss only if $\mathbb{E}_{q \sim \text{M}} \mathfrak{D}(q) > \mathbb{E}_{q \sim \text{U}} \mathfrak{D}(q)$, ensuring stability by restricting the adversarial loss to cases where the discriminator is likely to correctly identify the matches.

Our final loss for pretraining is defined as:
\begin{equation}
\mathcal{L}_{\text{final}} = \mathcal{L}_{\text{recon}} + w_{\text{var}} \, \mathcal{L}_{\text{var}} + w_{\text{cov}} \, \mathcal{L}_{\text{cov}} + w_{\text{adv}} \, \mathcal{L}_{\text{adv}}
\end{equation}

where $w_{\text{adv}}$, $w_{\text{var}}$ and $w_{\text{cov}}$ are the respective weighting factors for the adversarial, variance and covariance losses.

\section{Representations for Dictionary Retrieval}
\label{s:retrieval}

Sign dictionary retrieval is the task of identifying signs from a predefined database. This allows for quick lookup of related signs given a query sign which supports efficient dataset creation and promotes cross-linguistic understanding of signs \cite{schulder2024signs}.
The SignRep model can be used as a feature extractor for dictionary retrieval. Given a query sign video of length $L$, we apply a sliding window to extract segment representations $\mathbf{Z}^{\text{avg}} = \{ z_1^{\text{avg}}, z_2^{\text{avg}}, \dots, z_n^{\text{avg}}, \dots, z_N^{\text{avg}} \}$, where  $N = \left\lfloor \frac{L - T}{\text{stride}} \right\rfloor + 1$.
Since isolated signs typically begin and end in a resting pose, we use the output from 
\begin{equation}
    \gamma^{\text{act}}_n=\max(\mathcal{P}^{\{LH,\text{act}\}},\mathcal{P}^{\{RH,\text{act}\}})
\end{equation} to identify whether the signer is active. By checking hand activity, we compute a weighted average of the segment representations resulting in
\begin{equation}
    z^{\text{out}} = \frac{1}{\sum \gamma^{\text{act}}_n}\sum_{n=1}^N \gamma^{\text{act}}_n z^{avg}_n
\end{equation}
where $z^{\text{out}}$ is the representation for the query video.
Finally, we extract features for all dictionary videos and classify a query video by finding the closest match using cosine similarity of the normalized representations. \\

\noindent\textbf{Class Probability Distribution.} 
The retrieval approach enables the recognition of visually similar sign classes without additional training, which is useful for improving the accuracy of downstream tasks such as recognition. We construct a class distribution by computing the cosine similarity across all samples in the training dataset, capturing the visual proximity between them. From the resulting cosine similarity matrix, we compute inter-class similarities using the sign-sample labels to generate a matrix of shape $\mathbb{R}^{C \times C}$, where $C$ represents the number of classes. Applying a temperature-scaled softmax to each row of this matrix yields a probability distribution for each class relative to the others, forming a class probability distribution matrix $\phi\in\mathbb{R}^{C \times C}$, where each element of $\phi$ lies in the range $[0, 1]$.

We can incorporate the class probability distribution $\phi$ as a regularization term in the downstream recognition task using KL divergence. 
Given a sign video with target label $c$, we can compute the KL divergence between the predicted class distribution $\bar{y}$ (from the output classifier layer) and $\phi_c$ which is the pre-computed class probability distribution of $c$. This results in:
\begin{equation}
\mathcal{L}_{\phi} = \kappa \, \phi_c[c] \, \text{KL}(\phi_c \parallel \bar{y})
\end{equation}
where $\kappa$ is the weighting factor for the loss and $\text{KL}(\phi_c \parallel \bar{y})$ is the KL divergence. To emphasize the contribution of the target class distribution, we scale the KL divergence by $\phi_c[c]$, which is the value of the class probability for $\phi_c$ at $c$.

\section{Experiments}
\label{sec:exp}
We evaluate our model on sign recognition and retrieval using three datasets: ASL-Citizen \cite{desai2024asl}, WLASL2000 \cite{li2020word}, and NMFs-CSL \cite{hu2021global}. The ASL-Citizen dataset includes 2,731 isolated ASL signs recorded from a webcam. The WLASL2000 dataset, sourced from the web, contains 2,000 common ASL signs and presents challenges due to its noisy nature and limited samples per sign. The NMFs-CSL dataset consists of 1,067 Chinese Sign Language (CSL) signs, which require recognition of non-manual cues, such as facial expressions, to accurately identify signs.

\subsection{Evaluation Protocol}

We follow standard evaluation protocols for sign recognition, measuring top-1 and top-5 per-instance and per-class accuracy on WLASL and NMFs-CSL. For ASL-Citizen, we use the benchmark metrics specified for this dataset, including discounted cumulative gain (DCG), mean reciprocal rank (MRR), and top-1 and top-5 instance accuracy.

To evaluate our model as a frozen feature extractor, we apply a retrieval protocol based on the dictionary-based retrieval approach from \cite{desai2024asl}. We choose a retrieval approach over a linear evaluation protocol, as dictionary retrieval is a common and practical task in sign language processing. This method also offers a better evaluation of the generalization of the features to unseen datasets, which is particularly important given the evolving nature of sign languages.

\subsection{Experimental Setting}
\label{ss:exp}
\noindent\textbf{Pretraining.} We pretrain our model on the YouTube-SL-25 dataset \cite{tanzer2024youtube}, which consists of large-scale continuous sign videos from YouTube. We initialize the model with the base video Hiera-B architecture, pretrained on Kinetics using MAE. We randomly select 16 consecutive frames as inputs based on a co-articulated sign typically lasting for around 13 frames \cite{albanie2020bsl,pfister2013large,viitaniemi2014s}.
No changes are made to the original encoder architecture, while the decoder is replaced with our lightweight sign decoder.

\noindent\textbf{Recognition.} For downstream tasks, our approach leverages only the pretrained encoder, discarding the decoder and entirely eliminating the reliance on keypoint extraction. For recognition, we extend the pretrained model's input sequence to 64 frames. We preserve the pretrained weights by inflating the patch embeddings, which preserves the computational efficiency by avoiding increasing the number of tokens in the transformer, detailed in the Appendix. A linear classifier is applied to the pooled features from the encoder model to predict the sign classes.

\noindent\textbf{Retrieval.}  For dictionary retrieval evaluation, we follow the approach outlined in \cref{s:retrieval} using a stride of 2. For the label assignment, we compute the cosine similarity between the test video representation and those in the training dictionary. Each test query video is assigned the label of the training video that yields the highest similarity score.

Additional implementation details are provided in the Appendix.

\subsection{Evaluation on Sign Recognition}

\begin{table}[ht]
    \centering
    \begin{tabular}[width=\textwidth]{l c c c c}
        \toprule
        \multirow{2}{*}{} & \multicolumn{2}{c}{\textbf{Instance Acc.}} & \multicolumn{2}{c}{\textbf{Class Acc.}}\\
        \cmidrule{2-4} \cmidrule{5-5}
        \textbf{Method} & Top-1 & Top-5 &  Top-1  & Top-5 \\
        \midrule
          \rowcolor{gray!20} \multicolumn{5}{c}{\textbf{Skeleton-based}}\\
        \midrule
        ST-GCN \cite{yan2018spatial}      & 34.40 & 66.57 & 32.53 & 65.45 \\
        SignBERT \cite{hu2021signbert} & 39.40  &73.35 & 36.74 & 72.38 \\
        BEST  \cite{zhao2023best}  & 46.25 & 79.33 & 43.52 & 77.65 \\
        SignBERT+\cite{hu2023signbert+} & {\textbf{48.85}} & {\textbf{82.48}}& {\textbf{46.37}}& {\textbf{81.33}} \\
        \midrule
          \rowcolor{gray!20} \multicolumn{5}{c}{\textbf{Multi-modal}}\\
        \midrule
        BEST (+R) \cite{zhao2023best}        & 54.59 & 88.08 & 52.12 & 87.28 \\
        SignBERT(+R)  \cite{hu2021signbert}  & 54.69 & 87.49 & 52.08 & 86.93 \\
        SignBERT+(+R) \cite{hu2023signbert+} & 55.59 & 89.37 & 53.33  & 88.82 \\
        SAM ($5\xi$ )\cite{jiang2021skeleton}    & 58.73 & 91.46 & 55.93 & 90.94 \\
        SAM-v2 ($5\xi$ ) \cite{jiang2021sign}  & 59.39 & 91.48 & 56.63 & 90.89 \\
        NLA-SLR   \cite{zuo2023natural}    & 61.05 & 91.45 & 58.05 & 90.70 \\  
        NLA-SLR($3\xi$)  \cite{zuo2023natural}    & \textbf{61.26} & 91.77 & 58.31 & 90.91
        \\         
        StepNet (R+F) \cite{shen2024stepnet} & 61.17 & \textbf{91.94} & \textbf{58.43} & \textbf{91.43}  \\ 
        \midrule
          \rowcolor{gray!20} \multicolumn{5}{c}{\textbf{RGB-based}}\\
        \midrule
        I3D \cite{carreira2017quo}         & 32.48 & 57.31 & -     & -     \\
        I3D(BSL1K) \cite{albanie2020bsl}  & 46.82	& 79.36	& 44.72	& 78.47 \\
        StepNet  \cite{shen2024stepnet}  & 56.89 & 88.64 &54.54 & 87.97 \\
        \midrule
        \textbf{SignRep (Ours)} & \textbf{61.05} & \textbf{90.27} & \textbf{58.89} &  \textbf{89.44}\\
        \bottomrule
    \end{tabular}
    \caption{Comparison of downstream sign recognition results on WLASL2000. $\xi$ denotes a multi-crop inference.}
    \label{tab:wlasl_accuracy}
\end{table}

In \cref{tab:wlasl_accuracy}, our single-modality model demonstrates significant improvements over all existing single-modality approaches, achieving performance on par with complex multi-modal methods for WLASL2000. 
While ensembling with multi-modal approaches could further improve performance, we emphasize the importance of pretraining. Our Hiera encoder model with our pretraining, achieves these results without the extensive architectural modifications required by other methods. Instead of relying on additional modalities as input, our pretrained model has learned these features to create a robust sign representation model. Importantly, our pretraining strategy does not rely on any annotated sign data.

The effectiveness of our SignRep model on the NMFs-CSL dataset is further illustrated in \cref{tab:nmf_csl}, where we achieve a top-1 accuracy of 84.1\% and a top-5 accuracy of 98.8\%, outperforming all other methods, including multi-modal approaches. Notably, the closest single-modality comparison is the StepNet RGB model, which achieves 77.2\% in top-1 accuracy, nearly 7\% lower than our model. While StepNet focuses on developing specialized architecture for sign, we demonstrate our framework can effectively learn sign features from large-scale datasets. 

Our results on the ASL-Citizen dataset, shown in \cref{tab:aslcitizen}, further demonstrates the effectiveness of our model where it surpasses the previous baseline I3D model by 18\% in top-1 accuracy.

\begin{table}[t]
    \centering
    \begin{tabular}[width=\textwidth]{l c c}
        \toprule
        \textbf{Method} & \textbf{Top-1} & \textbf{Top-5}  \\
        \midrule
        I3D \cite{carreira2017quo}                & 64.4  & 88.0  \\
        TSM \cite{lin2019tsm}               & 64.5  & 88.7  \\
        Slowfast \cite{feichtenhofer2019slowfast}           & 66.3  & 86.6  \\
        GLE-Net \cite{hu2021global}            & 69.0  & 88.1  \\
        HMA ($\diamond$) \cite{hu2021hand}                & 75.6  & 95.3  \\
        StepNet  \cite{shen2024stepnet}          & 77.2  & 92.5 \\
        SignBERT (H+R) ($\diamond$) \cite{hu2021signbert}          & 78.4  & 97.3  \\
        BEST ($\diamond$) \cite{zhao2023best}              & 79.2  & 97.1  \\
        NLA-SLR  ($\diamond$) \cite{zuo2023natural}           & 83.4  & 98.3  \\
        StepNet (R+F) ($\diamond$) \cite{shen2024stepnet}         & 83.6  & 97.0   \\
        NLA-SLR ($\diamond$,$3\xi$) \cite{zuo2023natural}          & 83.7  & 98.5  \\
        \midrule
        \textbf{SignRep (Ours) }     & \textbf{84.1} & \textbf{98.8}     \\
        \bottomrule
    \end{tabular}
    \caption{Comparison of downstream sign recognition results on NMFs-CSL. $\diamond$ denotes methods using multi-modality/ensemble models and $\xi$ indicates a multi-crop inference.}
    \label{tab:nmf_csl}
\end{table}

\begin{table}[t]
    \centering
    \begin{tabular}[width=\textwidth]{l c c c c}
        \toprule
        \textbf{Model} & \textbf{DCG} & \textbf{MRR} &  \textbf{Rec@1}  & \textbf{Rec@5} \\
        \midrule
        ST-GCN* \cite{yan2018spatial}             & 76.37 & 69.97 & 59.52 & 82.68 \\
        I3D* \cite{carreira2017quo}                & 79.13 & 73.32 & 63.10 & 86.09 \\
        \midrule
        \textbf{SignRep (Ours)}      & \textbf{90.84}    & \textbf{88.05}     & \textbf{81.37}     & \textbf{96.11}    \\
        \bottomrule
    \end{tabular}
    \caption{Comparison of downstream sign recognition results on ASL-Citizen. * denotes results produced by \cite{desai2024asl}}.
    \label{tab:aslcitizen}
\vspace{-1.5em}
\end{table}

\subsection{Evaluation on Sign Dictionary Retrieval}

\begin{table*}[t]
    \centering
    \begin{tabular}[width=\textwidth]{l | c c c | c c c | c c c }
        \toprule
         { } & \multicolumn{3}{c|}{\textbf{ASL-Citizen}} & \multicolumn{3}{c|}{\textbf{WLASL2000}} & \multicolumn{3}{c}{\textbf{NMFs-CSL}}\\
        \cmidrule{2-10}
        {\textbf{Features}} & DCG  &  Rec@1  & Rec@5 & DCG  &  Rec@1  & Rec@5 & DCG  &  Rec@1  & Rec@5 \\
        \midrule
        HieraMAE-Kinetic      & 11.64  &  0.25  & 0.84   & 13.21  &  2.08  & 3.40 & 23.29  &  3.96  & 12.18 \\
        HieraMAE-YTSL   & 12.12  &  0.39  & 1.34   & 14.06  &  2.57  & 4.41 & 28.03  &  7.57  & 18.38 \\
        All Joint Angles    & 13.82  &  0.57  & 2.17   & 17.92  &  2.54  & 6.50 & 32.51  &  7.93  & 25.50 \\
        All Keypoints       & 14.29  &  0.98  & 2.74   & 19.37  &  3.16  & 8.23 & 36.64  &  12.26  & 30.99 \\
        Hand Keypoints  & 25.91  & 7.96   & 19.58  & 28.11  & 7.57   & 20.92 & 41.72  & 15.91   & 42.62 \\
        Hand Joint Angle      & 26.93  & 8.81   & 21.41  & 30.61  &  9.42  & 24.36 & 44.17  &  18.13  & 46.34 \\
        \midrule
        SignRep (avg)            & 61.40 & 37.47 & 68.77 & 53.57  &  26.16  & 60.98 & 79.51  &  58.48  & 91.85 \\
        SignRep (weighted)       & \textbf{71.21}  &  \textbf{49.95}  & \textbf{80.09} & \textbf{57.93}  &  \textbf{29.92}  & \textbf{67.41} & \textbf{83.05}  &  \textbf{63.04}  & \textbf{95.63} \\
        \bottomrule
    \end{tabular}
    \caption{Comparison of generalization results on sign retrieval with no downstream training applied to SignRep.}
    \label{tab:retrieval}
\vspace{-1.0em}
\end{table*}

To assess the effectiveness of our model as a feature extractor, we utilize a dictionary-based sign retrieval method that directly evaluates the quality of the learned representations without any fine-tuning. In \cref{tab:retrieval}, we compare our approach to the Hiera model pretrained with MAE on both the Kinetics and YT-SL datasets. We apply the same training settings outlined in \cref{ss:exp} for pretraining on YT-SL. Our results demonstrate that our proposed pretraining strategy significantly outperforms standard MAE, even when trained on YT-SL, by optimizing specifically for sign language representation, whereas MAE's pixel-reconstruction objective does not directly support learning the sign specific features needed for sign retrieval.

We also compare our model's retrieval performance with the use of raw 3D keypoints and joint angles averaged over time. As shown in \cref{tab:retrieval}, our method achieves a substantial performance increase, with top-1 retrieval scores more than tripling those based on hand joint angles. These results underscore the strength of our pretraining approach in capturing spatio-temporal features specific to sign language, establishing our method as a highly effective framework for sign representation learning. We also validate the importance of incorporating hand activity awareness, observing that weighted averaging based on hand activity improves retrieval performance compared to averaging.

\subsection{Ablation Studies}

\paragraph{Impact of Pretraining on Sign Retrieval.} In \cref{tab:retrieval_ab}, we examine the effects of various elements in our pretraining strategy on retrieval performance in WLASL2000. Our results show that combining all three types of sign priors that have angle, keypoint and distance yields the highest retrieval accuracy, underscoring the benefit of capturing multiple aspects of the sign features. We also analyze the role of masking during pretraining, removing it leads to a noticeable decline in retrieval performance, indicating that masking is essential for robust representation learning. While we observe lower reconstruction loss without masking, this does not translate to better retrieval scores, suggesting that precise reconstruction alone does not necessarily produce effective sign representations. Finally, the addition of both adversarial loss and variance-covariance regularization yields the strongest retrieval results, confirming that these feature regularization techniques improve the quality and robustness of learned sign representations.

\begin{table}[h]
    \centering
    \begin{tabular}[width=\textwidth]{c c c | c | c c | c}
        \toprule
        \multicolumn{3}{c|}{\textbf{Prior Comp.}} &  & \multicolumn{2}{c|}{\textbf{Reg.}} & \\
        \cmidrule{0-5}
        {angle} & {kpt} & {dist} & mask & var+cov & adv &  \textbf{DCG}  \\
        \midrule
        \checkmark &   &   & \checkmark &   &   & 45.1 \\
          & \checkmark &   & \checkmark &   &   & 32.9 \\
          &   & \checkmark & \checkmark &   &   & 47.2 \\
        \checkmark & \checkmark & \checkmark & \checkmark &   &   & 48.5 \\
        \checkmark & \checkmark & \checkmark &   &   &   & 46.3 \\
        \checkmark & \checkmark & \checkmark & \checkmark & \checkmark &   & 49.9 \\
        \checkmark & \checkmark & \checkmark &  \checkmark & \checkmark & \checkmark & \textbf{50.7} \\
        \bottomrule
    \end{tabular}
    \caption{Comparison of the impact of sign priors, masking and regularization on the impact of retrieval. Retrieval results are obtained on WLASL using a stride of 8.}
    \label{tab:retrieval_ab}
\vspace{-0.5em}
\end{table}

\paragraph{Impact of Pretraining for Recognition.} In \cref{tab:rec_ab}, we demonstrate that our pretrained model substantially outperforms both the MAE model pretrained on Kinetics and the model pretrained on YT-SL with pixel MAE. Additionally, the baseline Hiera model pretrained on Kinetics, achieving a top-1 accuracy of 51.5\%, surpasses prior keypoint-based masked learning methods such as SignBERT, BEST and SignBERT+, as shown in \cref{tab:wlasl_accuracy}. This result underscores the value of RGB input, revealing the limitations of using keypoints as a primary modality. Previous keypoint methods have required ensembling with RGB to remain competitive, illustrating the challenges of relying solely on skeletal information as an input modality for sign recognition. Finally, we observe that incorporating the sign class distribution loss further enhances recognition accuracy, improving top-1 accuracy from 59.9\% to 61.0\% with a $\kappa$ of 0.2.

\begin{table}[h]
    \centering
    \begin{tabular}[width=\textwidth]{l c c c c c}
        \toprule
        & &  \multicolumn{2}{c}{\textbf{Instance Acc.}} & \multicolumn{2}{c}{\textbf{Class Acc.}}\\
        \cmidrule{3-5} \cmidrule{6-6}
        {\textbf{Method}} & {$\kappa$} & top-1 & top-5 &  top-1  & top-5 \\
        \midrule
        MAE (Kinetic)  & -   & 51.5 & 83.4 & 48.4 & 82.0 \\
        MAE (YT-SL)    & -            & 57.2 & 87.4 & 54.6 & 86.3 \\
        SignRep       & 0.0             & 59.9 & 90.2 & 57.4 & 89.2  \\
        SignRep  & 0.1             & 60.4 & 90.1 & 57.8 & 89.1 \\
        SignRep    & 0.2             & \textbf{61.0} & \textbf{90.3} & \textbf{58.9} & \textbf{89.4} \\
        SignRep    & 0.5             & 59.8 & 89.4 & 57.7 & 88.8 \\
        \bottomrule
    \end{tabular}
    \caption{Comparisons of the impact of pretraining and weighting for the class distribution loss on WLASL.}
    \label{tab:rec_ab}
\vspace{-0.5em}
\end{table}

\section{Feature Extractor for Sign Translation}

\begin{table}[h]
    \centering
    \begin{tabular}[width=\textwidth]{l c c c c}
        \toprule
        \multirow{2}{*}{} & \multicolumn{2}{c}{\textbf{Phoenix14T}} & \multicolumn{2}{c}{\textbf{CSL-Daily}}\\
        \cmidrule{2-4} \cmidrule{5-5}
        {\textbf{Sign2GPT Backbone}}  & B-4 & R &  B-4 & R \\
        \midrule
        DinoV2(LoRA)  & 19.42 & \textbf{45.23} & 12.96 & 41.12  \\
        {SignRep(extracted)}&\textbf{20.38} &45.17 & \textbf{16.33} & \textbf{42.67} \\
        \bottomrule
    \end{tabular}
    \caption{Comparison of translation results using Sign2GPT \cite{wongsign2gpt} by replacing the learnable DinoV2 backbone with our extracted SignRep features. B-4 denotes BLEU4 and R denotes ROUGE.}
    \label{tab:translation}
\vspace{-1.0em}
\end{table}

\begin{table}[h]
    \centering
    \begin{tabular}[width=\textwidth]{l c c}
        \toprule
        \multirow{2}{*}{} & \multicolumn{2}{c}{\textbf{How2Sign}}\\
        \cmidrule{2-3}
        {\textbf{Tarr\'es's Backbone}}   & rBLEU & BLEU \\
        \midrule
        I3D(Supervised)  & 2.21 & {8.03}  \\
        {SignRep(Self-Supervised)} &\textbf{2.74} & \textbf{8.66}  \\
        \bottomrule
    \end{tabular}
    \caption{Comparison of translation results using Tarr\'es \cite{tarres2023sign}, replacing the I3D features with our SignRep features. }
    \label{tab:h2s}
\vspace{-1.0em}
\end{table}

In this section, we evaluate SignRep as a feature extractor for sign translation by integrating it into two open-source translation models: Sign2GPT \cite{wongsign2gpt} on RWTH-PHOENIX-Weather 2014T (Phoenix14T) \cite{camgoz2018neural} and CSL-Daily \cite{zhou2021improving}, as well as the approach from \cite{tarres2023sign} on How2Sign \cite{duarte2021how2sign}. We chose these models because they allow direct evaluation of SignRep as a frozen feature extractor without requiring sign-text pretraining or significant modifications to the architecture. We follow the same translation training hyperparameters as the original papers to ensure a fair comparison and isolate the impact of replacing the visual backbone.

For Phoenix14T and CSL-Daily, we omit the pseudo-gloss pretraining used in \cite{wongsign2gpt} to ensure a fair evaluation of SignRep's learned features without additional linguistic supervision. In \cite{wongsign2gpt}, they show that the DinoV2 backbone requires fine-tuning with LoRA to perform well on sign translation. Instead, we replace the original learnable DinoV2 backbone with pre-extracted SignRep features, obtained using our sliding window approach with a stride of 2.
This setup places ours at a disadvantage, as we remove the trainable backbone weights and data augmentation, relying solely on pre-extracted features. Despite this, SignRep achieves competitive translation performance, as shown in \cref{tab:translation}, performing comparably on Phoenix14T and achieving substantial gains on CSL-Daily.
By replacing the learnable DinoV2 model with pre-extracted SignRep features during training, we significantly reduce computational costs, making large-scale translation training more efficient and accessible without sacrificing performance. Our approach removes an additional 5.5GFLOP per frame required by the learnable DinoV2 model, which is important as translation models process tens to hundreds of frames. This leads to an 80\% reduction in GPU memory usage during training, lowering requirements from 60GB to 11GB with a batch size of 8. This reduction makes training on standard hardware more feasible, enabling broader accessibility for researchers working with sign translation models.

For How2Sign, in \cref{tab:h2s}, we replace the supervised I3D features with our self-supervised SignRep features on the translation framework from \cite{tarres2023sign} and observe performance improvements. Unlike I3D features, which are pretrained with supervision on a labeled sign recognition data, SignRep learns features purely through self-supervision, highlighting our framework's effectiveness for learning sign-specific features without supervision.

\section{Conclusions}

In this work, we introduced a scalable, self-supervised framework for sign representation learning without labeled sign datasets. Using masked autoencoding with sign priors, adversarial loss and feature regularizations, our approach enhances generalization while eliminating the need for skeletal keypoints, multimodal or multi-branch architectures in downstream tasks. Our single model achieves state-of-the-art performance on multiple sign recognition benchmarks with a simpler, more efficient design.
Beyond recognition, our representations demonstrate versatility in sign dictionary retrieval on unseen datasets and serve as an efficient feature extractor for existing sign translation systems, reducing computational costs while maintaining strong performance.
Our findings highlight the potential of our self-supervised learning framework for scalable sign modeling and encourage further research into practical applications of sign representations.

\section*{Acknowledgements}

This work was supported by SNSF project `SMILE II' (CRSII5 193686), European Union’s Horizon2020 programme (`EASIER' grant agreement 101016982) and the Innosuisse IICT Flagship (PFFS-21-47). This work reflects only the authors view and the Commission is not responsible for any use that may be made of the information it contains. Neither Necati Cihan Camgoz nor Meta were involved in the model training, evaluation, or use of the datasets. The authors also thank Maksym Ivashechkin for their assistance in providing the 3D pose estimation model.
{
    \small
    \bibliographystyle{ieeenat_fullname}
    \bibliography{main}
}

\clearpage
\setcounter{page}{1}
\setcounter{section}{0}
\renewcommand{\thesection}{A\arabic{section}} 
\maketitlesupplementary

\section{Human Pose Extraction}
\label{app:hpe}
To extract human pose features, we utilize angles derived from a human pose estimation model from \cite{ivashechkin2023improving}. We compute the bone lengths for all instances in the YouTube-SL-25 dataset (YT-SL) and select the median value as the standard bone length for each respective joint. This normalization ensures that all individuals are represented with the same body shape, thereby avoiding the leakage of person-specific features when converting angles into 3D keypoints.

\begin{figure*}[t]
    \centering
    \includegraphics[width=1.0\linewidth]{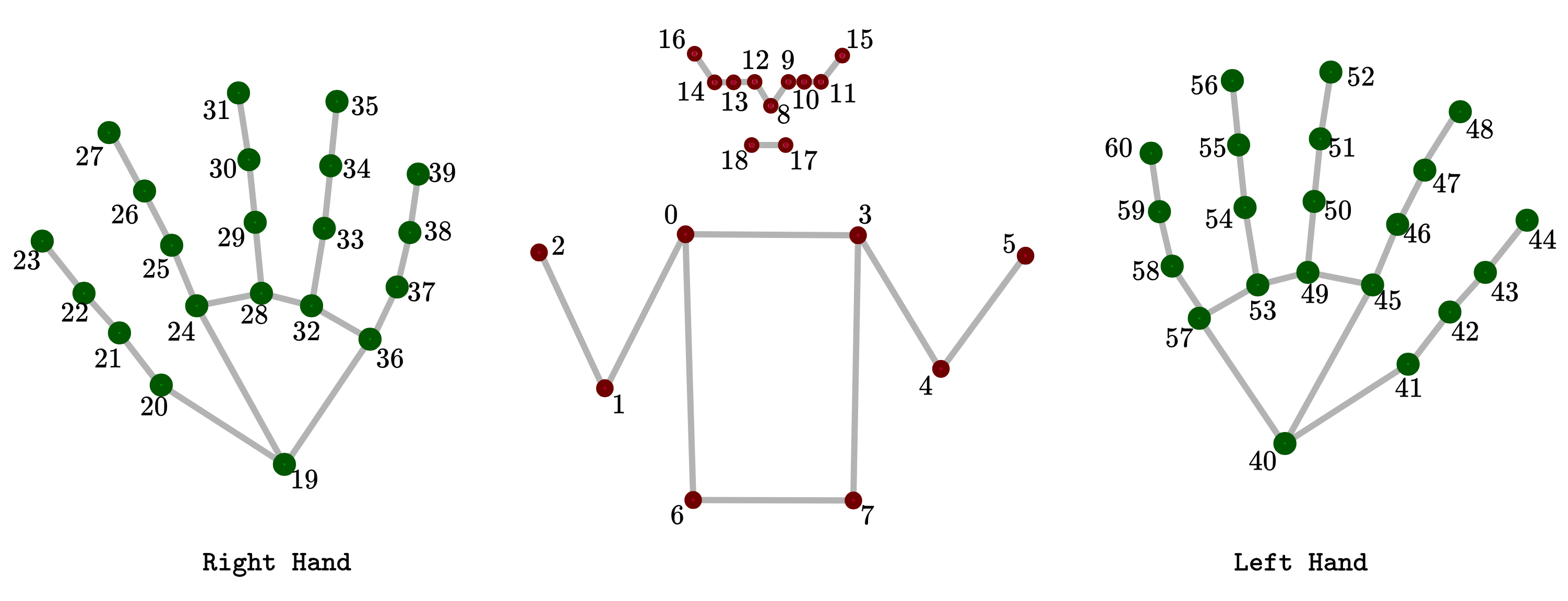}
    \caption{Visualization of 3D keypoint extracted. Numbers alongside the nodes represent the keypoint indices. For visualization purposes, we separate the left and right hand from the body.}
    \label{fig:kpts}
\end{figure*}

We visualize the resulting keypoints in \cref{fig:kpts}, separating the hands from the body for easier identification of indices. The left fingertips are defined using keypoint indices ${\{44,48,52,56,60\}}$. For the fingertip distance matrix, $\mathcal{P}^{\{b,d\}}$, these keypoints serve as the source, while indices ${\{40,41,44,45,48,49,52,53,56,57,60\}}$ are used as the destination for computing the distance matrix. Similarly, the same process is applied to the right hand using its respective keypoint indices.

For the hand-interaction distance prior, $\mathcal{P}^{\{b,d\}}$, we use the fingertip keypoint indices and the wrist keypoint ($\{40\}$ for left wrist and $\{19\}$ for right wrist) as the source. The destination includes the set of keypoint indices $\{0,3,6,7,10,\allowbreak 13,15,16,17,18,19,\allowbreak 23,27,31,35,39,40,\allowbreak 44,48,52,56,60\}$, which represent hands, face and body components. This matrix captures the distances between key positions involved in interactions between the hands and the rest of the body.

Human pose estimations often exhibit jitter across frames, which can affect temporal consistency. To mitigate this effect on the signer activity prior, $\mathcal{P}^{\{h,\text{act}\}}$, we determine whether a hand is inactive by checking two conditions: (1) its position is below the y-axis mean of keypoints $\{0,3,6,7\}$, and (2) the sum of the standard deviations across time for all 21 visible hand keypoints is less than 0.26. These criteria help identify inactive hands in the presence of keypoint jitter.

\section{Pretraining Dataset Processing}
\label{app:pdp}

For pretraining, we utilize the YT-SL dataset. We rely on pose estimations to ensure that a signer is present in each sequence, cropping the video to focus on the upper torso before resizing it to $256 \times 256$.

To prevent data leakage, since WLASL also contains YouTube videos, we ensure there is no overlap between the videos in the WLASL and YT-SL datasets. This is achieved by comparing the video IDs from WLASL with those in the YT-SL dataset, ensuring that no videos that are in the WLASL dataset are in our YT-SL pretraining data.

During pretraining, we randomly select 16 consecutive frames from each video. For each batch, we randomly select two sequences from the same video, ensuring that each batch contains a matching pair for the discriminator. 
These steps are then used to train the SignRep framework.

\section{Implementation Details}

As described in \cref{ss:exp}, we initialize the pretraining of our SignRep framework using the video Hiera Base model, pretrained with MAE on Kinetics. The output dimension $D$ is 768, with a drop path rate of 0.1. The sign decoder’s upsampler has a hidden dimension of 512 and the output dimension $D'$ is set to 384.

\paragraph{Pretraining.} During training, data augmentations include Planckian Jitter \cite{zini2022planckian}, random resized cropping from $256 \times 256$ to $224 \times 224$, Gaussian blur and grayscale conversion. The model is trained for 500,000 iterations with a batch of 20 and a masking ratio of 80\% on a single NVIDIA 3090 GPU. A warmup over the first 50,000 iterations gradually increases the learning rate to $1 \times 10^{-4}$ using the Adam optimizer \cite{kingma2014adam}, followed by cosine annealing decay. A layer-wise learning rate decay \cite{clark2020electra} is applied with a factor of 0.85.

In \cref{tab:hyp}, we list the hyperparameters used for the weighting of the loss functions during pretraining. Additionally, we apply a scaling factor $\psi$ to the target to balance the target values.

\begin{table}[h]
    \centering
    \begin{tabular}[width=\textwidth]{l c c}
        \toprule
        \textbf{Loss Components} & \textbf{weighting $w$} & \textbf{scale $\psi$}  \\
        \midrule
          \rowcolor{gray!20} \multicolumn{3}{c}{\textbf{Priors}}\\
        \midrule
        body angles ($w_{\mathcal{P}^{\{b,a\}}}$)    & 10.0  & 1.0  \\
        left hand angles ($w_{\mathcal{P}^{\{LH,a\}}}$)    & 10.0  & 1.0  \\
        right hand angles ($w_{\mathcal{P}^{\{RH,a\}}}$)    & 10.0  & 1.0  \\
        \midrule
        body kpt. ($w_{\mathcal{P}^{\{b,k\}}}$)    & 10.0  & 1.0  \\
        left hand kpt. ($w_{\mathcal{P}^{\{LH,k\}}}$)    & 10.0  & 2.0  \\
        right hand kpt. ($w_{\mathcal{P}^{\{RH,k\}}}$)    & 10.0  & 2.0  \\
        \midrule
        body dist. ($w_{\mathcal{P}^{\{b,d\}}}$)    & 20.0  & 1.0  \\
        left hand dist. ($w_{\mathcal{P}^{\{LH,d\}}}$)    & 20.0  & 4.0  \\
        right hand dist. ($w_{\mathcal{P}^{\{RH,d\}}}$)    & 20.0  & 4.0  \\
        \midrule
        signer activity ($w_{\mathcal{P}^{\{\text{act}\}}}$)    & 0.2  & -  \\
        \midrule
          \rowcolor{gray!20} \multicolumn{3}{c}{\textbf{Regularizations}}\\
        \midrule
        variance ($w_{\text{var}}$)    & 1.0  & -  \\
        covariance ($w_{\text{cov}}$)    & 0.004  & -  \\
        adversarial style  ($w_{\text{adv}}$)    & 2.0  & -  \\
        \bottomrule
    \end{tabular}
    \caption{Hyperparameters for weighting factors for the different loss components used during pretraining of SignRep.}
    \label{tab:hyp}
\end{table}

\paragraph{Downstream Recognition.} We use the same data augmentation as pretraining and apply cross-entropy loss with label smoothing of 0.1, with no patch masking applied, setting $\kappa$ to 0.2 for the class distribution loss. The model is trained with a batch size of 8 for 100 epochs, with 1000 iterations of warmup, followed by cosine annealing of the learning rate, with a max learning rate of $1 \times 10^{-4}$ using the Adam optimizer. The layer-wise learning rate decay factor is 0.85.

For the Adam optimizer, we utilize the AdamW version in Pytorch. We set the betas to $(0.9, 0.95)$ and use a weight decay of 0.5. To stabilize training, gradient clipping is applied with a maximum value of 1.0. During pretraining, the model is evaluated with retrieval on WLASL validation set every 25,000 iterations, the model achieving the best performance on the retrieval task using the WLASL validation set is selected for subsequent retrieval, recognition and translation tasks.

\paragraph{Downstream Translation.} For the downstream translation task, we use Phoenix14T, CSL-Daily and How2Sign. Phoenix14T \cite{camgoz2018neural} is a German Sign Language (DGS) dataset consisting of weather forecast broadcasts with aligned sign and text translations. CSL-Daily \cite{zhou2021improving} is a daily conversational Chinese Sign Language dataset recorded in a lab setting, covering various everyday interaction topics such as family life, shopping, travel and banking services. How2Sign \cite{duarte2021how2sign} is an American Sign Language (ASL) dataset that provides parallel signed video and text translations of instructional videos across a broad range of categories.

For a fair comparison, we use the open-source code from \cite{wongsign2gpt} for Phoenix14T and CSL-Daily and follow \cite{tarres2023sign} for How2Sign, applying the same hyperparameters specified in their respective papers. This ensures that improvements stem from our learned representations rather than differences in training configurations.

\section{Discriminator Setup}

In \cref{para:style}, the discriminator determines whether the output features $z^{\text{avg}}$ share the same style as a given style representation $z^{\text{style}}$. This process ensures that the representation encoder $f_{\text{enc}}$ learns style-agnostic representations, for robust and generalizable features.

The discriminator model is designed as a lightweight MLP-based architecture. To address the relatively small magnitude of the style representation values, $z^{\text{style}}$, we first scale these values by a factor of 100.0. The scaled style representation is then passed through a two-layer MLP with a hidden size of 768, which transforms it to match the dimensionality of $z^{\text{avg}}$. Layer normalization is applied after this transformation. Next, the transformed $z^{\text{style}}$ is concatenated with $z^{\text{avg}}$ and fed into a four-layer MLP with a hidden size of 768 and an output size of 1. This MLP is responsible for determining whether the representation of $z^{\text{avg}}$ aligns with the style $z^{\text{style}}$. Spectral normalization is incorporated into this final MLP to stabilize discriminator training. All linear layers, except the final linear layer, are followed by the GELU activation function.

Matched and unmatched style samples for training the discriminator are constructed from items within the batch. For each item in the batch, its matching styles are derived from its paired sample described in \cref{app:pdp}, while unmatched pairs are randomly selected from the remaining batch items. This setup ensures that the discriminator learns to distinguish between matching and non-matching styles effectively.

The discriminator is trained using binary cross entropy loss to predict 0's for unmatched styles and 1 for matched styles. We use a learning rate of $1\times 10^{-4}$, with a warm-up period of 50,000 iterations and cosine annealing decay. The Adam optimizer is used with betas $(0.5, 0.9)$ and a weight decay of $1\times10^{-3}$. 
An exponential moving average with an update momentum of 0.1 is used to compute the expected outputs of a matched style $\mathbb{E}_{q\sim \text{M}}\mathfrak{D}(q)$ and unmatched style $\mathbb{E}_{q\sim \text{U}}\mathfrak{D}(q)$. The discriminator is trained simultaneously with the SignRep representation model.

\section{Class Probability Distribution}

To create the class distribution $\phi$, we utilize the temperature-scaled distribution described in \cref{s:retrieval}. Our goal is to avoid excessively weak low-confidence probabilities and overly strong high-confidence probabilities, thereby achieving a smoother loss function $\mathcal{L}_{\phi}$.

For each class, we select a temperature $\tau$ such that the scaled distribution $\texttt{softmax}(\hat{\phi}_c / \tau)$ yields a maximum class probability as close as possible to, but still below, 0.5. Here, $\hat{\phi}_c$ represents the inter-class cosine similarity for class $c$. We determine the appropriate $\tau$ by iterating over values in the interval $[0.001, 0.1]$ and selecting the temperature that produces $\phi_c$ satisfying $\texttt{max}(\phi_c) < 0.5$ while being nearest to 0.5. We repeat this process for every class to obtain the final class distribution $\phi$.

\section{Inflated Patch Embeddings}

To accommodate a 64-frame input without increasing the number of tokens processed during the downstream recognition task, we employ \textit{inflated patch embeddings} as described in \cref{sec:exp}. This method preserves computational efficiency while capturing temporal relationships in the data. The pretraining is conducted on continuous sign data, whereas the downstream task involves isolated signs, which are temporally less dense. To address this discrepancy, we adapt the patch embeddings by inflating their temporal components, ensuring the preservation of temporal relations.

The original patch embeddings are defined with a kernel size of $(3, 7, 7)$, a stride of $(2, 4, 4)$, and padding of $(1, 3, 3)$. These parameters are updated to a kernel size of $(7, 7, 7)$, a stride of $(8, 4, 4)$, and padding of $(3, 3, 3)$. This adjustment allows for better modeling of the temporal relationships required for sign recognition without adding more patch tokens.

To ensure compatibility and preserve the pretrained weights, we employ a zero-initialization approach. The new kernel weights are first initialized to zero. Then, weights from the original patch embedding are mapped to the new kernels by transferring the weights from kernel indices $\{0, 1, 2\}$ to indices $\{1, 3, 5\}$ in the temporal dimension, respectively. 
This method ensures that the pretrained information is preserved during downstream initialization.

\section{Qualitative Retrieval}

\begin{figure*}[h]
\centering
\setlength{\tabcolsep}{0.1em}

{
\begin{tabular}{l c}
& \textbf{Top 3 Retrieved Video Segments on ASL-Citizen} \\
\\
\rotatebox[origin=c]{90}{\; \; \; \textbf{Ref.}}  &{\includegraphics[width=0.90\linewidth]{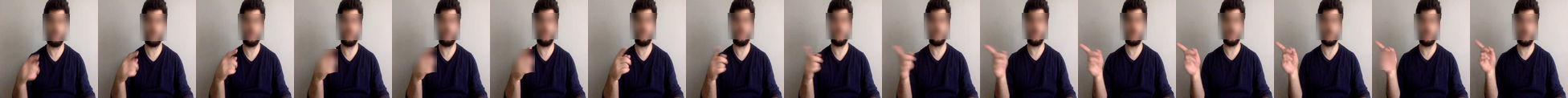}}  \\
\midrule
\rotatebox[origin=c]{90}{\; \; \; M1} & {\includegraphics[width=0.90\linewidth]{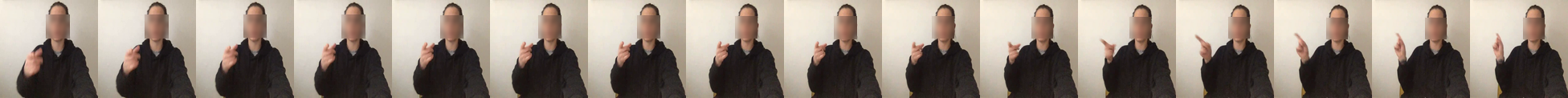}} \\
\rotatebox[origin=c]{90}{\; \; \; M2} & {\includegraphics[width=0.90\linewidth]{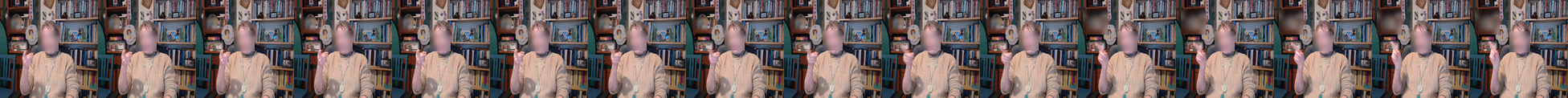}} \\
\rotatebox[origin=c]{90}{\; \; \; M3} & {\includegraphics[width=0.90\linewidth]{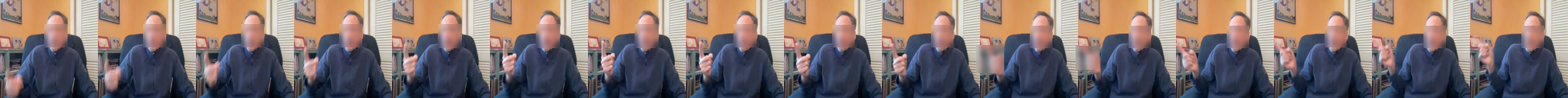}}  \\
\\
\\
\rotatebox[origin=c]{90}{\; \; \; \textbf{Ref.}}  &{\includegraphics[width=0.9\linewidth]{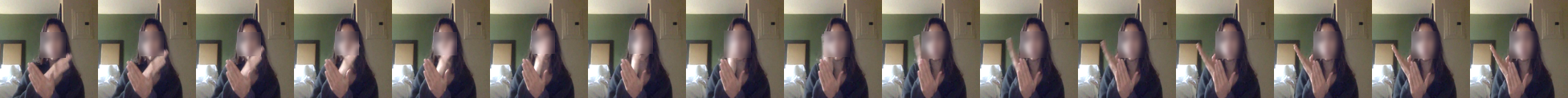}}  \\
\midrule
\rotatebox[origin=c]{90}{\; \; \; M1} & {\includegraphics[width=0.9\linewidth]{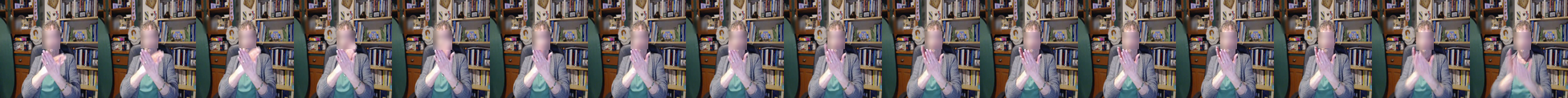}} \\
\rotatebox[origin=c]{90}{\; \; \; M2} & {\includegraphics[width=0.9\linewidth]{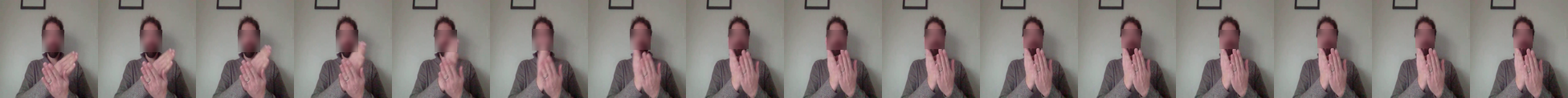}} \\
\rotatebox[origin=c]{90}{\; \; \; M3} & {\includegraphics[width=0.9\linewidth]{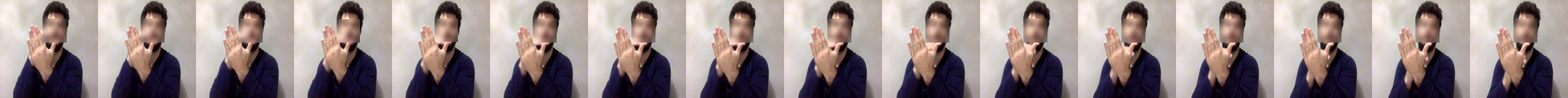}}  \\
\\
\\
\rotatebox[origin=c]{90}{\; \; \; \textbf{Ref.}}  &{\includegraphics[width=0.9\linewidth]{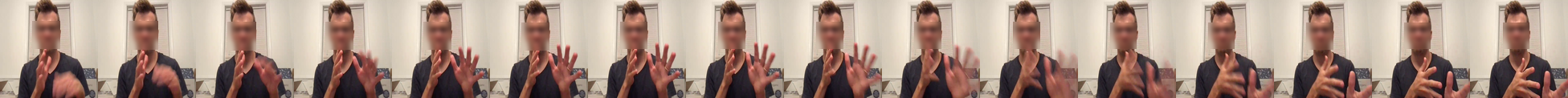}}  \\
\midrule
\rotatebox[origin=c]{90}{\; \; \; M1} & {\includegraphics[width=0.9\linewidth]{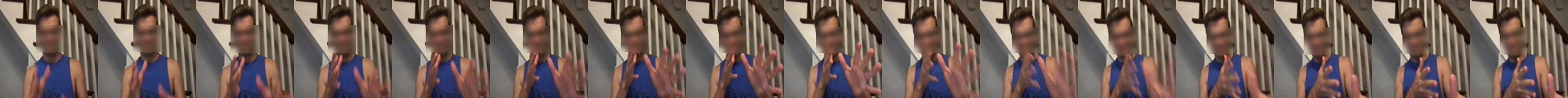}} \\
\rotatebox[origin=c]{90}{\; \; \; M2} & {\includegraphics[width=0.9\linewidth]{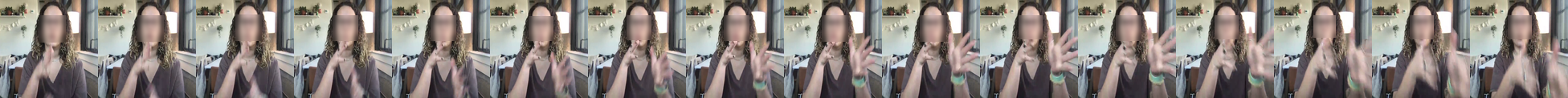}} \\
\rotatebox[origin=c]{90}{\; \; \; M3} & {\includegraphics[width=0.9\linewidth]{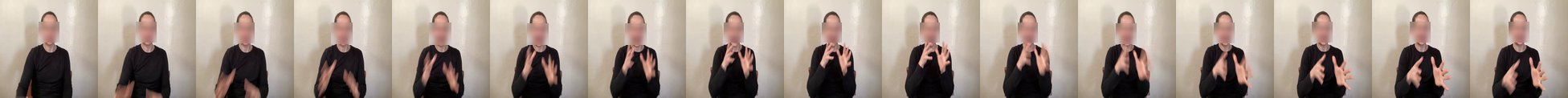}}  \\
\end{tabular}
}
\caption{Qualitative results for ASL-Citizen for retrieval based on features extracted from the pretrained SignRep. Given the reference sequence (Ref.), the Top-3 most similar videos are retrieved based on the cosine similarity of the output representations. M1 denotes the closest match, M2 is the second closest match and M3 is the third closest match.}
\label{fig:qual-asl}
\end{figure*}

\begin{figure*}[h]
\centering
\setlength{\tabcolsep}{0.1em}

{
\begin{tabular}{l c}
& \textbf{Top 3 Retrieved Video Segments on NMFs-CSL} \\
\\
\rotatebox[origin=c]{90}{\; \; \; \textbf{Ref.}}  &{\includegraphics[width=0.90\linewidth]{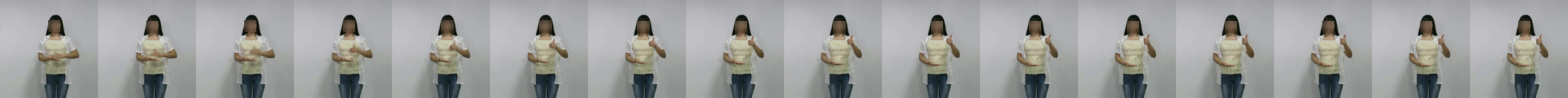}}  \\
\midrule
\rotatebox[origin=c]{90}{\; \; \; M1} & {\includegraphics[width=0.90\linewidth]{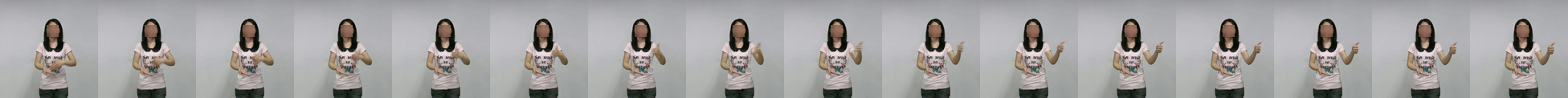}} \\
\rotatebox[origin=c]{90}{\; \; \; M2} & {\includegraphics[width=0.90\linewidth]{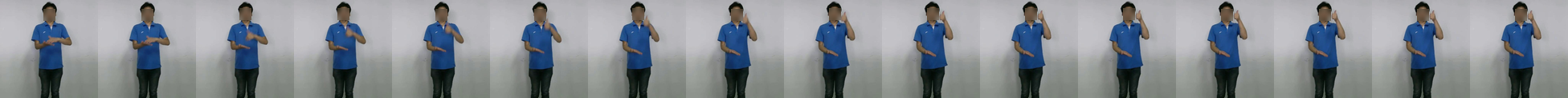}} \\
\rotatebox[origin=c]{90}{\; \; \; M3} & {\includegraphics[width=0.90\linewidth]{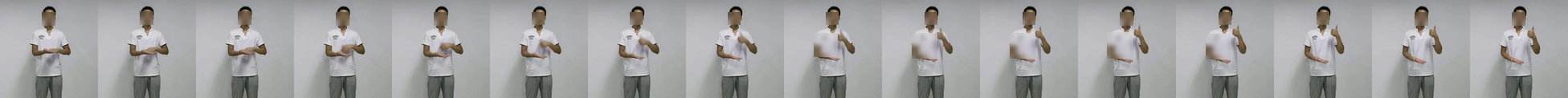}}  \\
\\
\\
\rotatebox[origin=c]{90}{\; \; \; \textbf{Ref.}}  &{\includegraphics[width=0.9\linewidth]{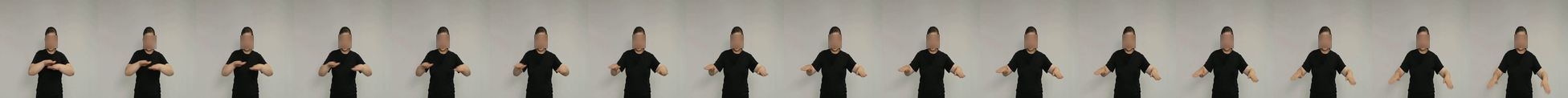}}  \\
\midrule
\rotatebox[origin=c]{90}{\; \; \; M1} & {\includegraphics[width=0.9\linewidth]{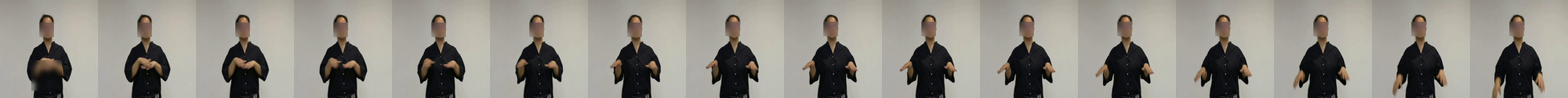}} \\
\rotatebox[origin=c]{90}{\; \; \; M2} & {\includegraphics[width=0.9\linewidth]{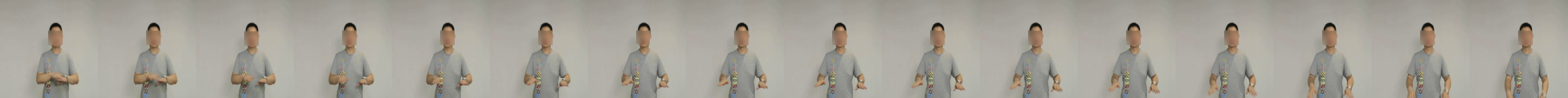}} \\
\rotatebox[origin=c]{90}{\; \; \; M3} & {\includegraphics[width=0.9\linewidth]{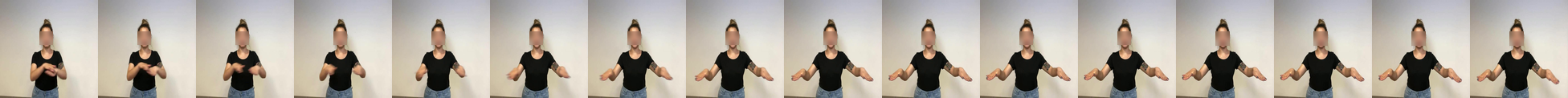}}  \\
\\
\\
\rotatebox[origin=c]{90}{\; \; \; \textbf{Ref.}}  &{\includegraphics[width=0.9\linewidth]{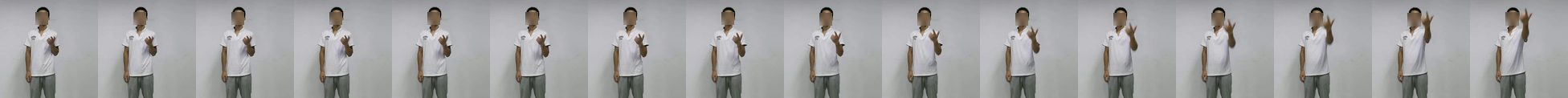}}  \\
\midrule
\rotatebox[origin=c]{90}{\; \; \; M1} & {\includegraphics[width=0.9\linewidth]{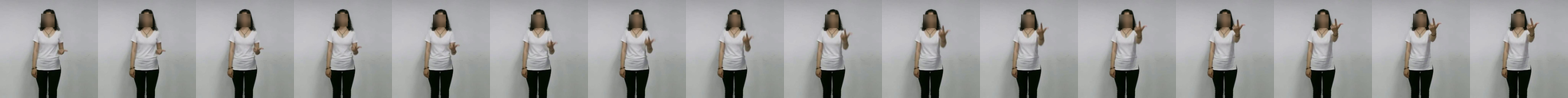}} \\
\rotatebox[origin=c]{90}{\; \; \; M2} & {\includegraphics[width=0.9\linewidth]{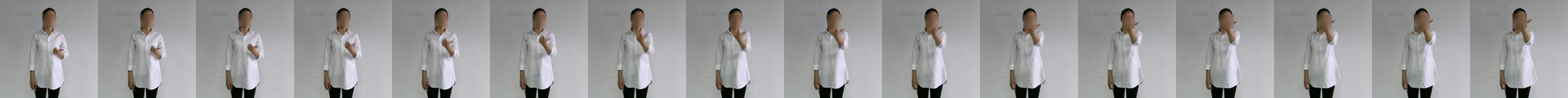}} \\
\rotatebox[origin=c]{90}{\; \; \; M3} & {\includegraphics[width=0.9\linewidth]{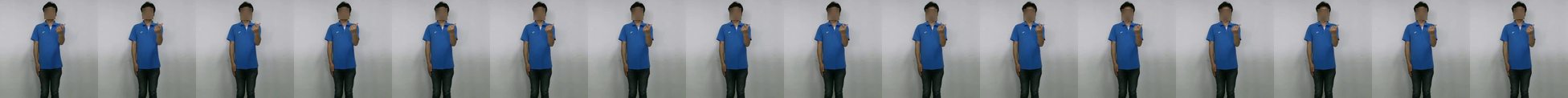}}  \\
\end{tabular}
}
\caption{Qualitative results for NMFs-CSL for retrieval based on features extracted from the pretrained SignRep. Given the reference sequence (Ref.), the Top-3 most similar videos are retrieved based on the cosine similarity of the output representations. M1 denotes the closest match, M2 is the second closest match and M3 is the third closest match.}
\label{fig:qual-nmf}
\end{figure*}

\begin{figure*}[h]
\centering
\setlength{\tabcolsep}{0.1em}

{
\begin{tabular}{l c}
& \textbf{Top 3 Retrieved Video Segments on WLASL} \\
\\
\rotatebox[origin=c]{90}{\; \; \; \textbf{Ref.}}  &{\includegraphics[width=0.90\linewidth]{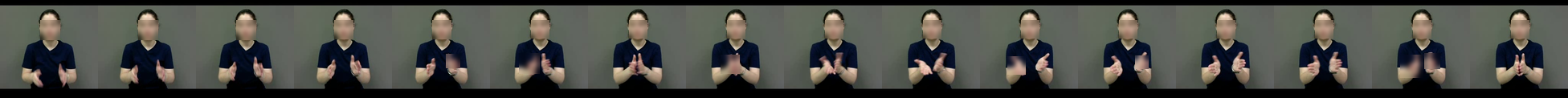}}  \\
\midrule
\rotatebox[origin=c]{90}{\; \; \; M1} & {\includegraphics[width=0.90\linewidth]{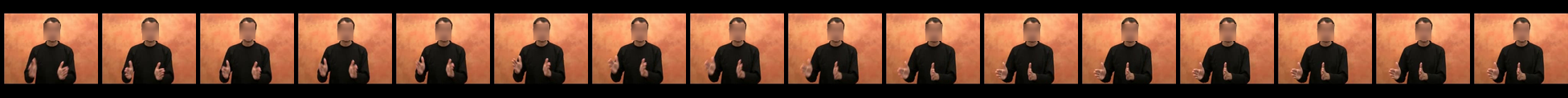}} \\
\rotatebox[origin=c]{90}{\; \; \; M2} & {\includegraphics[width=0.90\linewidth]{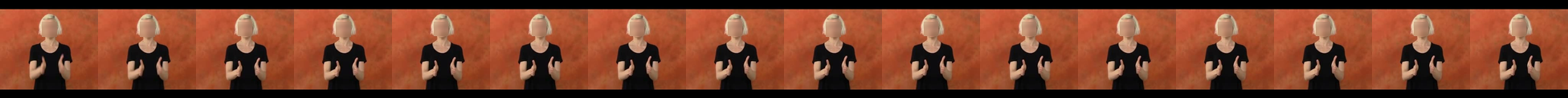}} \\
\rotatebox[origin=c]{90}{\; \; \; M3} & {\includegraphics[width=0.90\linewidth]{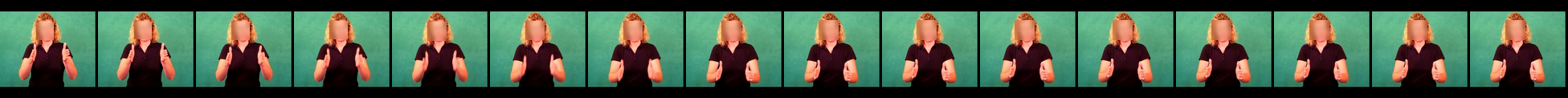}}  \\
\\
\\
\rotatebox[origin=c]{90}{\; \; \; \textbf{Ref.}}  &{\includegraphics[width=0.9\linewidth]{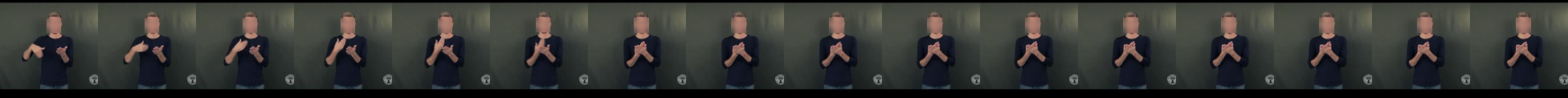}}  \\
\midrule
\rotatebox[origin=c]{90}{\; \; \; M1} & {\includegraphics[width=0.9\linewidth]{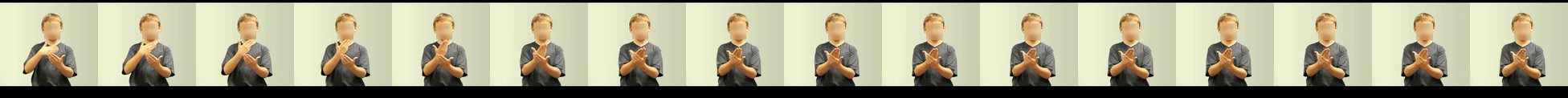}} \\
\rotatebox[origin=c]{90}{\; \; \; M2} & {\includegraphics[width=0.9\linewidth]{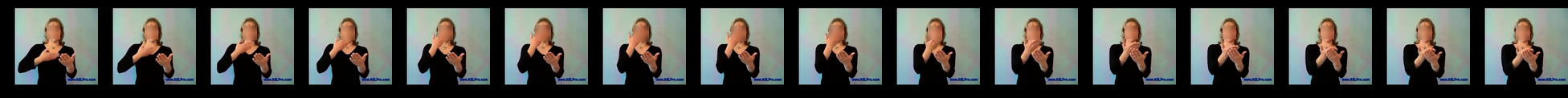}} \\
\rotatebox[origin=c]{90}{\; \; \; M3} & {\includegraphics[width=0.9\linewidth]{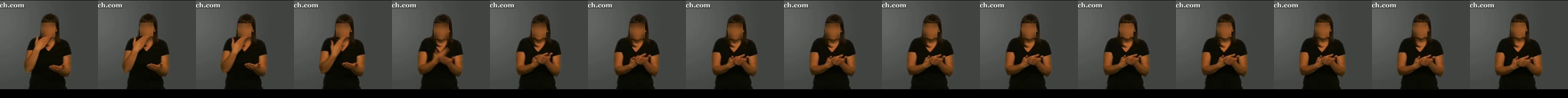}}  \\
\\
\\
\rotatebox[origin=c]{90}{\; \; \; \textbf{Ref.}}  &{\includegraphics[width=0.9\linewidth]{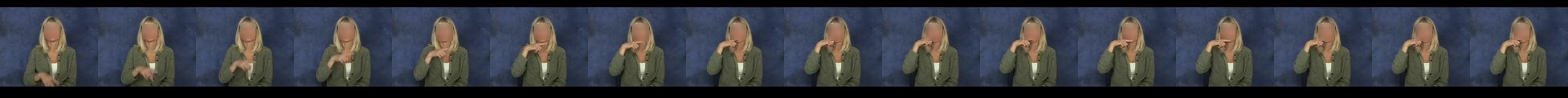}}  \\
\midrule
\rotatebox[origin=c]{90}{\; \; \; M1} & {\includegraphics[width=0.9\linewidth]{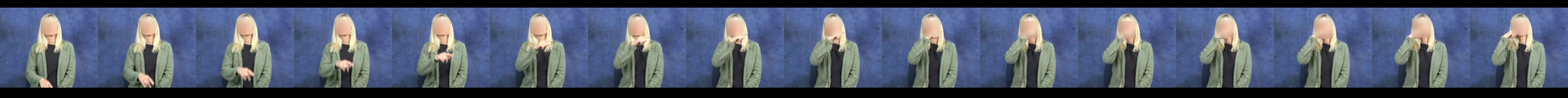}} \\
\rotatebox[origin=c]{90}{\; \; \; M2} & {\includegraphics[width=0.9\linewidth]{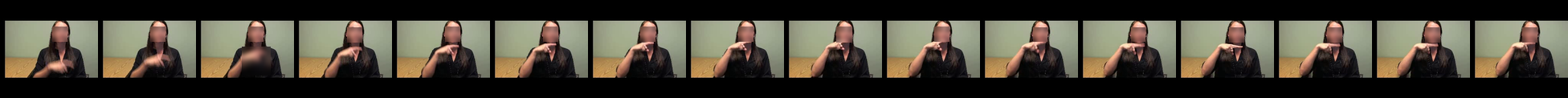}} \\
\rotatebox[origin=c]{90}{\; \; \; M3} & {\includegraphics[width=0.9\linewidth]{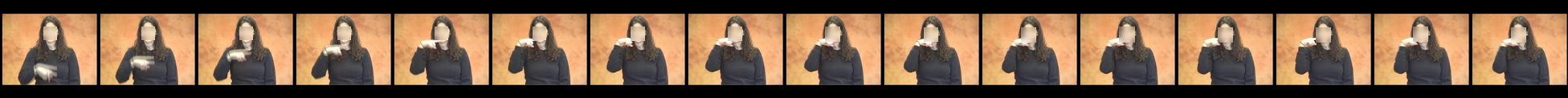}}  \\
\end{tabular}
}
\caption{Qualitative results for WLASL for retrieval based on features extracted from the pretrained SignRep. Given the reference sequence (Ref.), the Top-3 most similar videos are retrieved based on the cosine similarity of the output representations. M1 denotes the closest match, M2 is the second closest match and M3 is the third closest match.}
\label{fig:qual-wlasl}
\end{figure*}

We show qualitative results of the pretrained SignRep model on the three downstream recognition datasets, ASL-Citizen in \cref{fig:qual-asl}, NMFs-CSL in \cref{fig:qual-nmf} and WLASL in \cref{fig:qual-wlasl}. We note that the retrieved results are generated using the pretrained model, which has neither been fine-tuned on the downstream recognition task nor exposed to the downstream video dataset during pretraining. We display the top-3 closest retrieved video segments for randomly selected reference video segments with active signers. The results show that the model effectively retrieves segments with similar hand shapes, poses and motions, highlighting its ability to capture meaningful sign-related features during pretraining.

\section{Limitations}

Our model is pretrained on Youtube-SL-25, which carries inherent limitations in terms of signer diversity, language distribution and skin tone representation. These factors may affect the quality and generalizability of the learned representations.
Additionally, our method focuses solely on manual sign features, leaving room for future improvements by incorporating non-manual components such as facial expressions and mouthing patterns. 
While our approach eliminates the need for keypoints during downstream tasks, the pretraining process still relies on keypoint-based supervision, which may be affected by low-quality detections. To mitigate this, we leverage a human pose estimation model specialized for sign language \cite{ivashechkin2023improving}. Furthermore, we filter out keypoints with confidence scores below 50\% and mask missing keypoints in the loss function. These adjustments are advantageous over methods relying on keypoints as inputs.

Our model learns individual sign representations using a 16-frame window. Future work could explore extending this to longer temporal windows. However, doing so would require careful modifications to prevent excessive computational overhead, as increasing the number of frames also increases token complexity. Alternatively, our model can serve as a lightweight feature extractor for learning inter-sign relationships and long-range temporal dependencies in a more efficient manner.

\end{document}